\documentclass[runningheads,a4paper]{llncs}

\usepackage{amssymb}

\usepackage{pgfplots}
\pgfplotsset{compat=1.14}
\usepackage{xcolor}
\usepackage{tikz}

\definecolor{bblue}{HTML}{4F81BD}
\definecolor{rred}{HTML}{C0504D}
\definecolor{ggreen}{HTML}{9BBB59}
\definecolor{ppurple}{HTML}{9F4C7C}

\usepackage{graphicx}
\usepackage{url}
\newcommand{\keywords}[1]{\par\addvspace\baselineskip
\noindent\keywordname\enspace\ignorespaces#1}

\begin{document}

\mainmatter  

\title{Shimon the Robot Film Composer and DeepScore}

\titlerunning{Shimon and DeepScore}

\author{Richard Savery\and Gil Weinberg }


\authorrunning{Savery and Weinberg}

\institute{Georgia Institute of Technology \\ \email{\{rsavery3, gilw\} @gatech.edu}}

\maketitle

\begin{abstract}
Composing for a film requires developing an understanding of the film, it’s characters and the film aesthetic choices made by the director. We propose using existing visual analysis systems as a core technology for film music generation. We extract film features including main characters and their emotions to develop a computer understanding of the film's narrative arc. This arc is combined with visually analyzed director aesthetic choices including pacing and levels of movement. Two systems are presented, the first using a robotic film composer and marimbist to generate film scores in real-time performance. The second software-based system builds on the results from the robot film composer to create narrative driven film scores.

\keywords{Film composition, algorithmic composition, visual analysis, artificial creativity, deep learning, generative music}
\end{abstract}

\section{Introduction}
Film composition requires a ``connection with the film" and a deep knowledge of the film's characters\cite{Karlin2004}. The narrative arc is often tied together through key themes developed around characters; as Buhler notes ``motifs are rigidly bound to action in film" \cite{Buhler2000}. Multiple authors have studied the link between musical themes and on-screen action \cite{Jarvis2015}\cite{Simmons2017}. Likewise, Neumeyer explores the relation of audio and visual, describing multiple visual and aural codes that link music and screen\cite{Neumeyer2015}. 

This deeply established practice emphasizing the relation between music, narrative and visuals, contrasts with existing computer film musical generative systems which focus on small form pieces, and do not include any video analysis. By including visual analysis and the film itself as intrinsic to the creation process, generative systems can begin to address the inherent challenges and opportunity presented in film composition. Analysis of film visuals also allow for a range of new approaches to generative music, while encouraging new musical and creative outcomes. 

This research began by exploring what it means for a robot composer to watch a film and compose based on this analysis. With lessons learned from this implementation we were able to prototype a new software-based approach to film generation using visual analysis. This paper will explore the design of both systems focusing on how visuals are tied to generative processes. The first system \textit{The Space Between Fragility Curves} utilizes Shimon, a real-time robotic composer and marimbist that watches and composes for the film. This system acted as a catalyst to the work developed for \textit{DeepScore}. The second system \textit{DeepScore} is off-line and uses deep learning for visual analysis and musical generation. In both systems multiple visual analysis tools are used to extract low level video features that are then converted to meta level analysis and used to generate character and environment based film scores.

\section{The Space Between Fragility Curves}
This project was built around the concept of Shimon acting like a traditional silent film composer. The system was created for a video art piece called \textit{The Space Between Fragility Curves} directed by Janet Biggs, set at the Mars Desert Research Station in Utah. A single video channel version premiered on the 17th of May 2018, at the 17º Festival Internacional de la Imagen in Manizales, Colombia. The two channel version premiered on the 14th of November, May 2018 at the Museo de la Ciencia y el Cosmos, in Tenerife, Canary Islands. The final film includes footage of Shimon interspersed amongst the footage from the Mars Research station. 

\begin{figure}
\centering
\includegraphics[height=6.2cm]{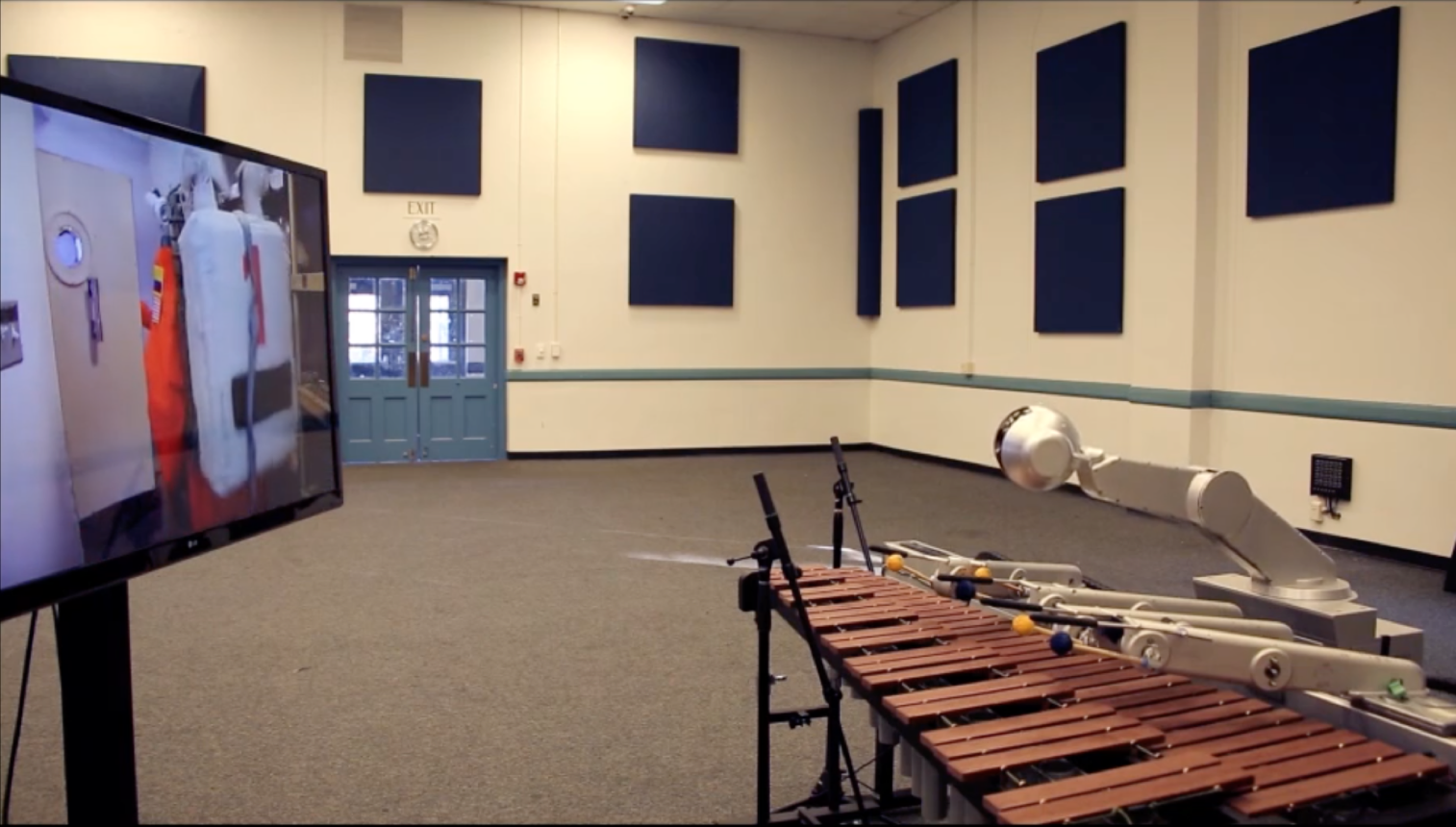}
\caption{Shimon Practicing The Space Between Fragility Curves \textit{(Photo by Janet Biggs)}}
\label{fig:example}
\end{figure}

\subsection{Shimon}
Shimon (see Figure 1.) is a robotic marimba player, developed by the Robotic Musicianship Group at Georgia Tech, led by Gil Weinberg \cite{Hoffman2010}. Shimon's body is comprised of four arms, each with two solenoid activators striking mallets on the marimba. Shimon has toured worldwide and is used as a platform for many novel and creative musical outcomes. 

\subsection{Visual Analysis}
With a set film given at the beginning of the project, visual analysis focused on extracting key elements that a composer may track within this specific film. Custom analysis tools were built in MaxMSP's Jitter, reading a JSON file generated by Microsoft's Video Insights. The JSON file included identified faces and their time and location on screen. It also included objects and a location analysis, defining if the scene was indoors or outdoors as well as the surrounding landscape. Jitter was used to extract director aesthetic choices, defined as conscious film choices that set the tone, style and pace of film. These tracked aesthetic choices were panning, zoom, similarity between camera angle changes, character movement and coloration. Parameters were then used to create a meta analysis of the overall pacing. 

\subsection{Musical Generation}
A musical arc was set by the director, dividing the film into four minutes of character and object driven musical generation, with two segments given independent algorithmic processes. At the beginning of each run, four melodies are generated using a Markov model trained on melodies from Kubrick film scores. A Markov chain is a probability based model that bases future choices on past events. In this case we referred to three past events, using a third generation Markov Model for pitch and a separate fifth generation model for rhythm, both trained on the same data. Two melodies are assigned to characters, with the other two melodies set for indoor and outdoor scenes. Melodies are then used throughout the film, blending between each one dependent on what is occurring on screen. Melodies are varied based on movement of the chosen characters on screen, their position and external surroundings. 

The first of the separate sections was centered inside an air chamber with director requesting a claustrophobic soundtrack. For this scene an embodied approach was used with Shimon. Shimon's design encourages the use of semitones, as both can be hit without the arms moving. Chords were then built around a rule set featuring chords built on these intervals. The second section was the conclusion of the film, which uses Euclidean rhythms \cite{Toussaint2005}, commonly used in algorithmic composition. The scene features one of the main characters riding an ATV across the desert. The number of notes per cycle is set based upon the movement of the ATV and position on screen. 

\subsection{Lessons Learned}
For this system we did not conduct any user studies. We considered it a prototype to generate ideas for a complete system. As previously mentioned the film is premiering in the Canary Islands and has multiple other showings lined up around the world indicating a certain level of success. Comments from the film director also demonstrated the importance of the link between visuals, ``I love how clear it is that Shimon is making choices from the footage" \footnote{Biggs, Janet. ``Re: Demos" Message to Richard Savery, 26 October, 2017. Email}. Follow up emails also suggested that the generative material was effective however the overall musical structure could be improved ``I watched the demos again and am super happy with the opening segment. The only other note that I would give for the second half is to increase the rhythm and tonal (tune) quality" \footnote{Biggs, Janet. ``Re: Demos" Message to Richard Savery, 6 November, 2017. Email}. 

After positive feedback from the director and informal viewings we reviewed the key concepts that were beginning to emerge. Extracting director aesthetic choices such as movement on screen and panning allowed an instant level of feedback that helped align the score with visuals. To some degree this naturally creates musical arcs matching the film's arc, however with only this information the music is always supporting the visuals and never complementing or adding new elements to the film. Like-wise character based motives were very successful from our small viewing sessions yet without intelligently changing these based on the character they also fell into a directly supporting role. Most significantly we came to believe that there was no reason for future systems to work in real-time. Shimon composing a new version for the film through each viewing provides a level of novelty but in reality the film will always be set beforehand and live film scoring is a vastly different process to the usual work flow of a film composer.

\section{DeepScore}
In contrast to Shimon acting as a real-time composer, \textit{DeepScore} was created to be used off-line for a variety of films. This encouraged a significantly different approach to visual analysis parameters and methods used for musical generation. Where Shimon as a film composer focused on building a real-time system for one film, \textit{DeepScore} aims to instead use more general tools to enable composition for multiple films. 

\subsection{DeepScore Backgound}

\subsubsection{Film Score Composition}
A successful film score should serve three primary functions, tonally matching the film, supporting and complementing the film and entering and exiting when appropriate\cite[p.10]{Hill2017}. From as early as 1911 film music (then composed for silent films) was embracing Wagner's concept of the leitmotif \cite[p. 70]{Buhler2000}. The leitmotif in film is a melodic gesture or idea associated with a character or film component \cite[p. 42]{Buhler2000}. While not the only way to compose for film, leitmotif's use has remained widespread most prominently by John Williams, but also by many other composers\cite{bribitzer-stull_2015}. 

\subsubsection{Deep Learning for Music}
DeepScore's musical generation and visual analysis rely on deep learning, a subfield of machine learning that uses multiple layers to abstract data into new representations. Deep learning has recently seen widespread adoption in many fields, driven by advances in hardware, datasets and benchmarks, and algorithmic advances \cite[p.20]{Chollet2007}. In this paper we use Recurrent Neural Networks (RNNs) for music generation. RNNs are a class of neural networks that are used for processing sequential data. An RNN typically consists of one or more nodes (operating units) which feed their outputs or hidden states back into their inputs. In this way, they can handle sequences of variable length, and allow previously seen data points in a sequence to influence the processing of new data points and are thought to have a sort of memory. Standard RNNs suffer from a variety of issues which make them difficult to train, and so most applications of RNNs today use one of two variations known as Long Short Term Memory (LSTM) RNNs and Gated Recurrent Unit (GRU) RNNs. In each of these variations, the standard recurrent node is replaced with one which parameterizes a memory mechanism explicitly. An LSTM recurrent has three gates: the input gate, cell gate, and output gate, which learn what information to retain and what information to release. RNNs have been used widely in music generation. Magenta (part of Google's Brain Team) have successfully used RNNs for multiple systems to create novel melodies \cite{45871}\cite{45935}.

\subsubsection{Deep Learning for Visuals}
For visual processing we primarily rely on Convolutional Neural Networks (CNN). A CNN is a neural network which uses one or more convolutional layers in their architecture and specializes them for the processing of spatial data. A convolutional layer applies an N-Dimensional convolution operation (or filtering) over it's input. CNNs have been shown to have the ability to learn spatially invariant representations of objects in images, and have been very successfully applied to image classification and object recognition \cite[p. 322]{GoodfellowIanBengioYoshuaCourville2016}. CNNs have also been used for musical applications, notably in WaveNet a model that generates raw audio waveforms\cite{wavenet}. WaveNet's model was itself based on a system that was designed around image generation, PixelCNN \cite{pixelcnn}.

\subsubsection{Film Choice - Paris, je t'aime}
A general tool for all films is far beyond the scope of DeepScore. We chose to limit the system to use on films where using leitmotifs for characters would be appropriate and emotional narrative arcs are present. Technical limitations of our visual system also restricted the system to films with primarily human main characters.For the purpose of this paper all examples shown will be from the 2006 film,\textit{ Paris, je t'aime} \cite{Coen2006} a collection of 18 vignettes. The examples will use the vignette \textit{Tuileries} directed by Joel and Ethan Coen. This film was chosen as it focuses on three main characters who experience a range of emotions. 

\subsection{DeepScore System Outline}
\textit{DeepScore} was written in Python, and uses Keras running on top of Tensorflow. Visual analysis is central to the system with the meta-data created through the analysis used throughout. Figure 2 demonstrates the flow of information through the system. Two separate components analyze the visuals, one using deep learning and the other computer vision. These two visual analysis units combine to create visual meta-data. The lower level data from the visual analysis is also kept and referenced by the musical generation components.  Melodies and chords are independently created and annotated. With visual data extracted, the generated chords and melodies are separately queried to find those best fit to the characters in the analyzed film. After chords and melodies are chosen they are combined together through a separate process, using a rule-based system. These melodies and chord progressions are then placed throughout the film. After placement they are then altered with changes in tempo, chord and melody variations and counter melodies added, dependent on the visual analysis.

\begin{figure}
\centering
\includegraphics[height=9.2cm]{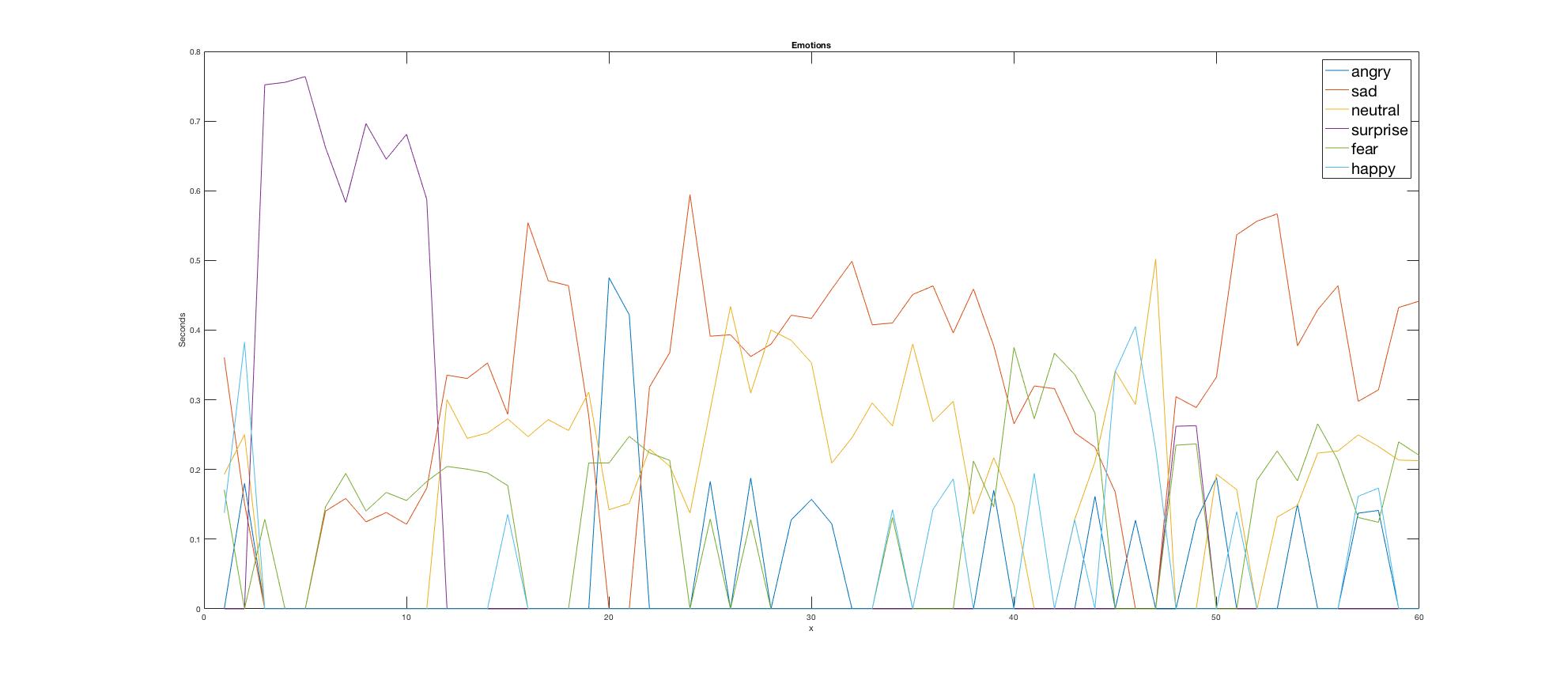}
\caption{DeepScore System Outline}
\label{fig:example}
\end{figure}

\subsection{Visual Analysis}
The primary visual element tracked by \textit{DeepScore} are the main characters and their emotions throughout the film. The three main characters are identified by which face appears most on screen. Emotions are tracked throughout the film using an implementation of an existing CNN \cite{facecnn} and the Facial Expression Recognition 2013 (FER-2013) emotion dataset\cite{Kaggle}. FER-2013 was chosen as it is one of the largest recent databases and contains almost 36,000 faces tagged with seven expressions, happy, angry, sad, neutral, fear, disgust or surprise. Each frame with a face recognized is given a percentage level of each emotion (see Figure 3). Figure 4 shows the first 60 seconds of emotional analysis of Paris, je t'aime

\begin{figure}
\centering
\includegraphics[height=4.2cm]{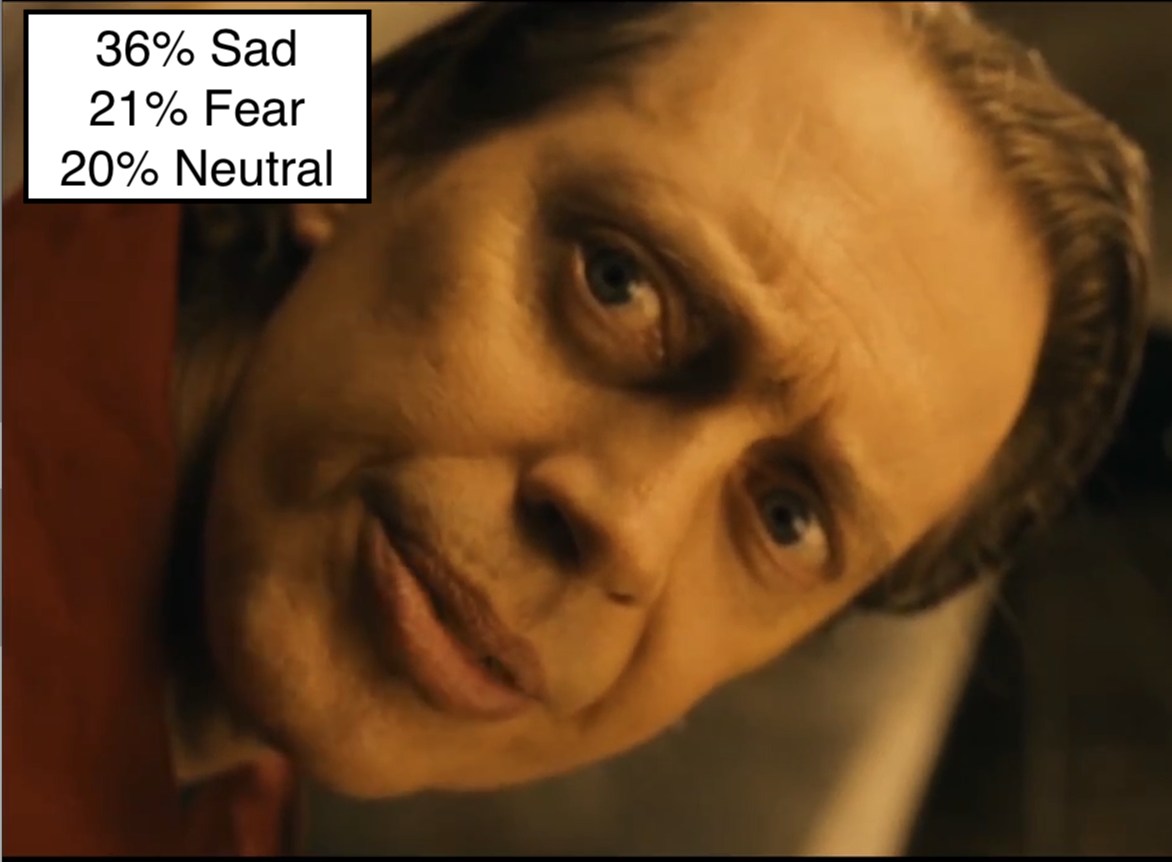}
\caption{Emotion Classification from \textit{Paris, je t'aime}}
\label{fig:example}
\end{figure}

Emotion was chosen for multiple reasons, at a simplistic level emotion is often considered a key component of both film and music. As previously discussed music should support the tone and complement the film, both characteristics relying on understanding the emotional content of a scene. Figure 4 demonstrates the emotional arc of the first sixty seconds.

\begin{figure}
\centering
\includegraphics[height=6.2cm]{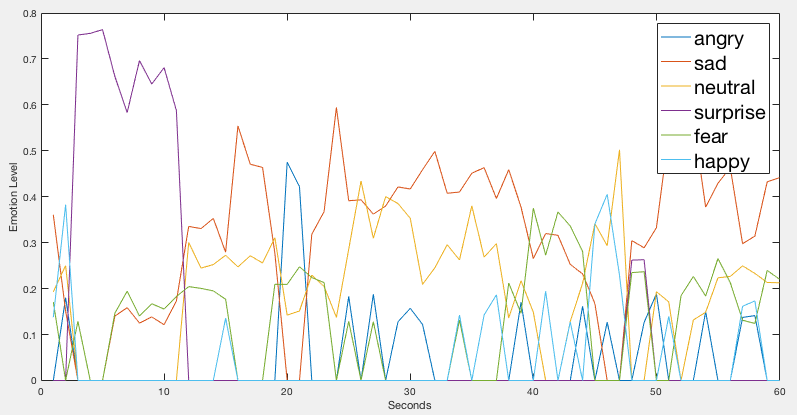}
\caption{Graph of Emotions}
\label{fig:example}
\end{figure}

In addition to emotions, the previously created custom analysis system for director aesthetics was used for DeepScore. The emotions were then combined with the director aesthetics into higher level annotations. These key points dictated when the musical mood would change and set the transition points from moving between characters. 

\subsection{Musical Generation}
\subsubsection{Chord Generation}
Chord progressions are generated using a character recurrent neural network, based on Karpathy’s[7] char RNN model and using the Band-in-a-box data set. The data set contains 2,846 jazz chord progressions. This type of RNN is often used to generate text. Character RNN is a recurrent neural network architecture which generates the next step in a sequence conditioned on only the previous step. Running DeepScore creates 1000 chord progressions, all transposed to C. These are then annotated with consonance level and variation, both between 0 and 1. Consonance consists of how closely the chords align to chords built off the scales of  either C Major, or C Minor. For example in C Major chords such as D minor 7 and E Minor 7 are given a 0 for consonance, D7 is given a 0.5 and Db Minor would be given a 1.0. The variation level refers to how many different chords are within a progression. 

\subsubsection{Melody Generation}
Melodic ideas are also created by an RNN in this case an LSTM and uses the Nottingham dataset, a collection of 1000 folk tunes. We tested multiple datasets, but found the folk melodies in this dataset worked best for their ability to be easily rearranged post creation and still retain their melodic identity.  Each created melody is 8 bars long. Melodies are then annotated based on rhythmic and harmonic content using custom software written in python, combined with MeloSpy from the Jazzomat Research Project. Table 1 shows each extracted parameter. Annotation is based on principles on a survey of research into the impacts of musical factors on emotion \cite{Juslin2010}.

\begin{table}
\begin{center}
    \begin{tabular}{ | l | p{6cm} |}
    \hline
    Musical Feature & Parameters Extracted \\ \hline
    Harmony & Consonance, Complexity \\ 
    Pitch & Variation \\ 
    Interval & Size, direction and consonance  \\ 
    Rhythm & Regularity, Variation \\ 
    Tempo & Speed range\\ 
    Contour & Ascending, Descending, Variation \\ 
    Chords & Consonance, Variation \\ \hline    
    
    \end{tabular}
    \caption{Musical Parameters for Emotional Variations}
    \end{center}
    \end{table}

\subsection{Adding Music to Image}
With chords and melodies annotated the system then uses the visual analysis to choose an appropriate melody for each character. This melody then becomes the leitmotif for the character. Starting with the character most present throughout the film three melodies and three chord progressions are chosen that align with the emotional arc of the main character. Two other characters are then assigned melodies primarily choosing features that align with their emotional arc, while contrasting that of the main character. 

At this point the chord progression and melody have been independently created and chosen. To combine them a separate process is used that alters notes in the melody, while maintaining the chord progression. Notes that do not fit within each chord are first identified and then shifted using a rule based system. This system uses melodic contour to guide decisions, aiming to maintain the contour characteristics originally extracted. Figure 5 shows two melodies and chord progressions after being combined. Both were chosen by the system for one character.

Main themes are then placed across the film, with motifs chosen by the dominant character in each section. In the absence of a character a variation of the main characters theme is played. After melodies are placed, their tempo is calculated based on the length of the scene and the character's emotion. Dependent on these features either a 2, 4 or 8 measure variation is created. 

\begin{figure}
\centering
\includegraphics[width=11cm]{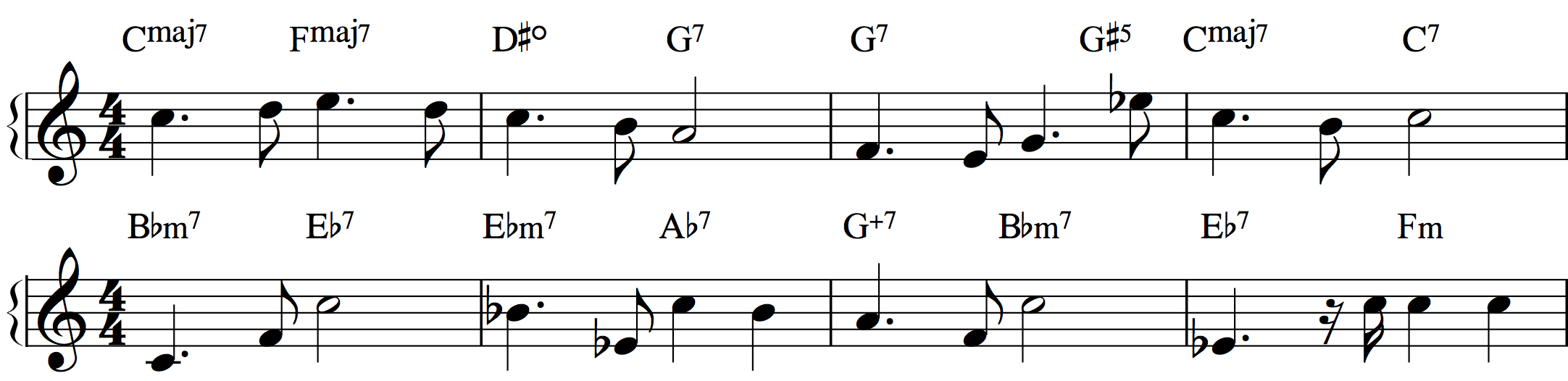}
\caption{Two Alternate Melodies Created for the Main Character}
\label{fig:example}
\end{figure}

\subsection{Counter Melodies and Reharmonization}
Referring back to the visual analysis meta-data each section of the film is then given a counter melody or reharmonization dependent on the emotional characteristic in the scene. These melodies are developed by a rule based system using the same parameters presented in table 1. All emotions are mapped to different variations based on a survey of studies in the relation between music and emotional expression\cite[p.384-387]{Juslin2010}. In addition to those parameters counter melodies use volume variation and articulations (staccatos and tenutos). The interactions between characters are also considered, such as when one character is happy while another is fearful. 

\subsection{Output}
The system's final output is MIDI file and a chord progression. This file doesn't contain instrument designations, but is divided by octaves into melody, counter-melody, bass-line and chordal accompaniment. In the current iteration this is then orchestrated by a human. The aim in future work is to include orchestration in the system.

\section{Evaluation}
\subsection{Process and Participants}
For the evaluation we created three, one minute long clips of music generated by the system. The first clip used \textit{DeepScore} to generate material, but did so based on emotional keywords representative of the short film as a whole, and not at a frame by frame level. This eliminated the immediate visual link to the film and was used to gauge a reaction to the generated music. Considerable work was done to ensure that this first clip was indistinguishable in terms of quality and processes to that of music generated purely by visuals. The other two clips used visual analysis as described in this paper on two different scenes. The three clips were randomly ordered for each participant. For each clip participants were asked three questions and responded with a rating between zero and ten. These questions were: \textit{How well does the music fit the film's tone, how well does the music complement the on-screen action and what rating would you give the music considered separately to the film?} They were then given an option to add any general comments on the score. After answering questions on the generated material they were presented a brief overview of a potential software program implementing visual analysis. Participants response's were anonymous and they were not given any background information on how the music was created, or that it was computer generated.

We surveyed five film composers and five film creators. The film composers were all professional composers having a collective composing experience of over 700 publicly distributed films. Film creators were directors, editors or writers or often a combination of the three. Film creators had a combined experience of over 100 publicly distributed films. While only a small sample group, we chose to focus our evaluation on leading industry participants to gauge just how effective the system is to those with the greatest insight into the area. 

\subsection{Quantitative Findings}
Figure 6 and 7 present the results from the survey asking about tone, if the music complements the scene, and the general rating of the music. The versions that included visual analysis were rated better across all categories, despite using the same generative system. Composers rated the music of the non-visual analysis particularly low. There was also a consistent gap between the ratings of the film creators and film composers.

\begin{figure}
\begin{tikzpicture}
    \begin{axis}[
        width  = 0.85*\textwidth,
        height = 5cm,
        major x tick style = transparent,
        ybar=2*\pgflinewidth,
        bar width=14pt,
        ymajorgrids = true,
        ylabel = {Rating},
        symbolic x coords={Tone,Complementary,Rating},
        xtick = data,
        scaled y ticks = false,
        enlarge x limits=0.25,
        ymin=0,
        ymax=10,
        legend style={
			cells={anchor=east},
			legend pos=outer north east,
        }
    ]
        \addplot[style={bblue,fill=bblue,mark=none}]
            coordinates {(Tone, 6.0) (Complementary,6.0) (Rating,6.5)};
        \addplot[style={ppurple,fill=ppurple,mark=none}]
            coordinates {(Tone, 3.33) (Complementary,4.3) (Rating,2.5)};
        \legend{Film Creator, Film Composer}
    \end{axis}
\end{tikzpicture}
\caption{Ratings for Music Generated with Keywords} \label{fig:M1}
\end{figure}
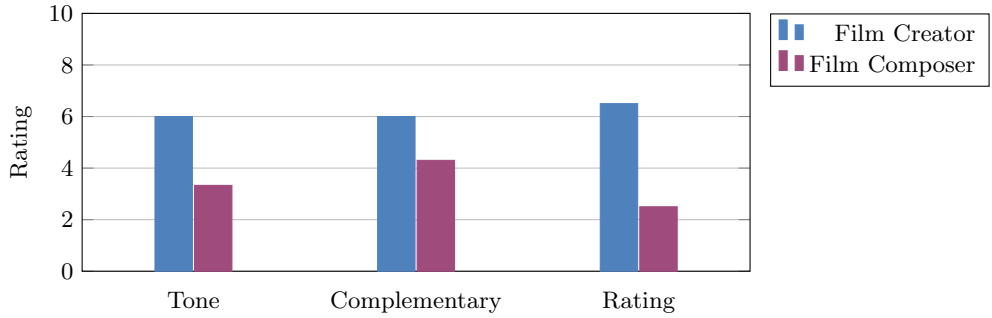

\begin{figure}
\begin{tikzpicture}
    \begin{axis}[
        width  = 0.85*\textwidth,
        height = 5cm,
        major x tick style = transparent,
        ybar=2*\pgflinewidth,
        bar width=14pt,
        ymajorgrids = true,
        ylabel = {Rating},
        symbolic x coords={Tone,Complementary,Rating},
        xtick = data,
        scaled y ticks = false,
        enlarge x limits=0.25,
        ymin=0,
        ymax=10,
        legend style={
			cells={anchor=east},
			legend pos=outer north east,
        }
    ]
        \addplot[style={bblue,fill=bblue,mark=none}]
            coordinates {(Tone, 7.75) (Complementary,7.35) (Rating,7.875)};
        \addplot[style={ppurple,fill=ppurple,mark=none}]
            coordinates {(Tone, 5.25) (Complementary,5.875) (Rating,4.625)};
        \legend{Film Creator, Film Composer}
    \end{axis}
\end{tikzpicture}
\caption{Ratings for Music Generated with Visual Analysis } \label{fig:M1}
\end{figure}
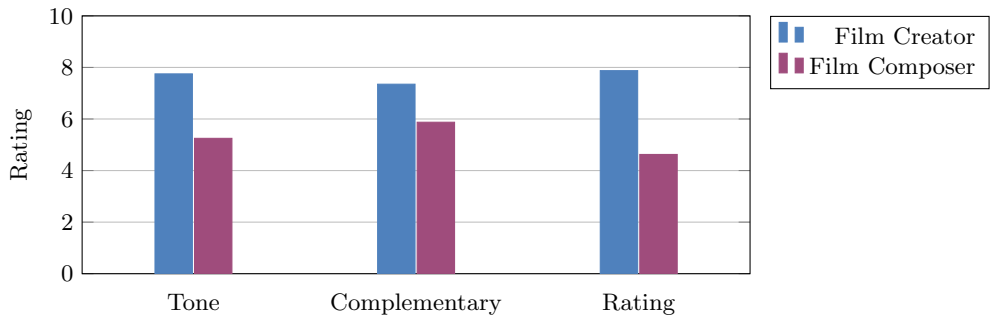

\subsection{Qualitative Findings}
From the general comments to the scores we had many unexpected insights and found significant variation between creator's and the composer's responses. For the version generated with keywords and no visual analysis one director noted: \textit{``The music as such is good to listen. But I don't think it is in sync with the actions on the screen".} Another director described, \textit{``Simple score but is not taking me emotionally anywhere"}. The composers also described the lack of relation to the emotions on screen, \textit{I think that the music is in the right direction but lacking the subtlety of action and emotion on the screen.} and it \textit{``doesn't seem to be relevant to what's happening on the screen in terms of tonality"}.  Participants had no previous insight into our analysis of character's emotions in other scenes. 

As expected from the quantitative ratings, comments were generally more positive for the versions using \textit{DeepScore's} visual analysis. One director noted that the music was \textit{``appropriate for the scene"}, a composer described \textit{``it generally fits very well"} and another composer mentioned \textit{``the music does well with changing action and emotion".} Two composers did notice that the transitions that occurred as the scene changed could be \textit{``a bit jarring"} and \textit{``the sudden change is too extreme".}

A key finding stemmed from the impact the music was having on the interpretation of the film. One director said that they \textit{``feel like the music is very much directing my feelings and understanding of what is going on.  If music had different feel the action could be interrupted differently"}. This related to one of the most common observations from composers, that the score \textit{``plays the action, although it didn't add to more than what we already see on screen"} with another composer describing they would have led the scene in a different direction \textit{``to either something more comical or menacing".}

From the qualitative findings we drew three main conclusions on how the system operated. By closely following the emotions, changes can be too significant and draw attention to moments on screen that are already very clear to the audience. This leads on to a larger problem that the current mapping of emotions to the musical generative system one dimensionally scores the film. It is currently never able to add new elements and contrasting emotional analysis to the scene. Finally, the systems music and mapping to the screen dictates a mood onto the scene. This in itself isn't necessarily a negative or positive but restricts the applications of \textit{DeepScore}. 
 
\subsection{Potential Applications}
\begin{figure}
\centering
\includegraphics[width=11cm]{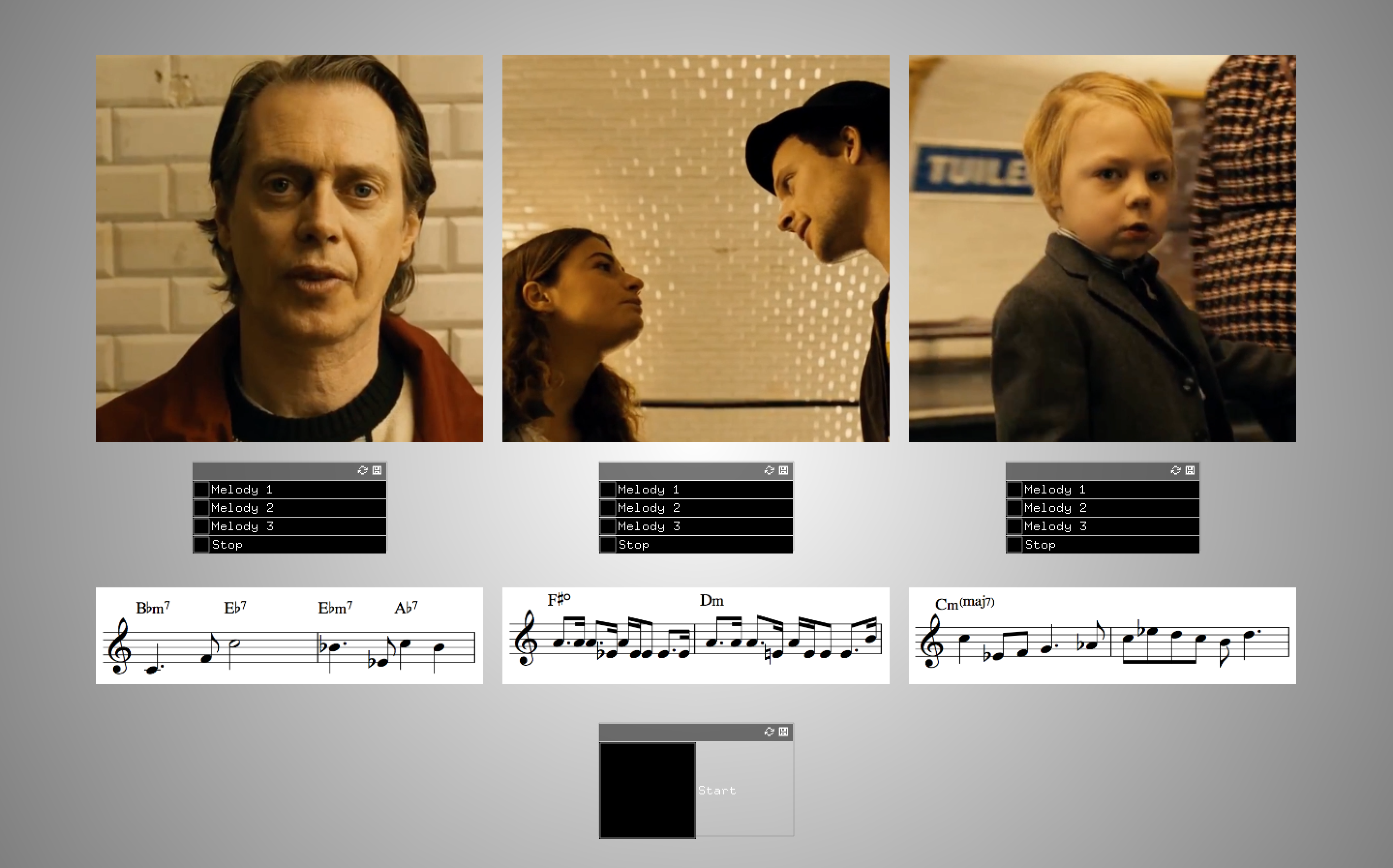}
\caption{Program Demonstrating Potential Application}
\label{fig:example}
\end{figure}
To conclude the survey we presented a short video demonstrating a potential implementation of the system for general use (see Figure 8). This tool showed three characters from the film, and allowed the user to choose a melody for each one (from a list of three generated melodies) and then hear the film with those melodies. One composer outright dismissed using any assistance to compose, believing composing should be a personal humanistic activity. The other composers were more open minded to the tool, two in particular proposed being able to use the system to augment their own work and create new variations. Another composer mentioned they would use it to create quick demos to test ideas, but not to compose an entire project. Directors were generally interested in using the tool however the main concerns were the tool had to be easy to use and simple to modify, but most importantly better than the temp tracks.

\section{Conclusion}
In this paper we have described a software based film composer, built from successes and lessons learned while creating a robot film composer. There have been consistent links to visual analysis improving the relation of generated materials to the film. Linking to visuals not only improved the connection to the film, but also improved the rating and perception of the music's quality.  General responses to the system showed that an emotional dialogue between score and visuals is central to connecting to the film. In future iterations this dialogue needs to become multiple dimensional, whereby emotions on the screen can be complemented in ways other than a direct musical response. Although only briefly analyzed, we also contend there are many applications for using visual analysis as a tool in film music generation for both composers and film creators.

\subsection{Acknowledgements}
Many thanks to Shannon Hwu, Matthew Kaufer and Zach Kondack.

\bibliographystyle{abbrv}
\bibliography{My_Collection.bib}

\end{document}